\documentclass[twoside]{article}

\usepackage[accepted]{aistats2023}
\usepackage[round]{natbib}

\usepackage[utf8]{inputenc} 
\usepackage[T1]{fontenc}    
\usepackage{hyperref}       
\usepackage{url}            
\usepackage{booktabs}       
\usepackage{amsfonts}       
\usepackage{nicefrac}       
\usepackage{microtype}      
\usepackage{xcolor}         

\usepackage{graphicx}
\usepackage{comment}
\usepackage{bm}
\usepackage{mathtools}
\usepackage{amssymb}
\usepackage{enumitem} 
\usepackage{amsthm}
\usepackage{booktabs}
\usepackage{xfrac}
\usepackage{algorithm}
\usepackage{algorithmic}
\usepackage{wrapfig}
\newtheorem{theorem}{Theorem}
\theoremstyle{definition}

\newtheorem{lemma}{Lemma}

\begin{document}

%
\runningtitle{Adversarial Random Forests}

%
\runningauthor{Watson, Blesch, Kapar, \& Wright}

\twocolumn[

\aistatstitle{Adversarial Random Forests for\\Density Estimation and Generative Modeling}

\aistatsauthor{David S. Watson \And Kristin Blesch}

\aistatsaddress{King's College London \And 
Leibniz Institute for Prevention\\ Research and Epidemiology – BIPS,\\ University of Bremen} 

\aistatsauthor{Jan Kapar \And Marvin N. Wright}

\aistatsaddress{
Leibniz Institute for Prevention\\ Research and Epidemiology – BIPS,\\ University of Bremen \And 
Leibniz Institute for Prevention\\ Research and Epidemiology – BIPS,\\ University of Bremen,\\University of Copenhagen} 
]

\begin{abstract}
  We propose methods for density estimation and data synthesis using a novel form of unsupervised random forests. Inspired by generative adversarial networks, we implement a recursive procedure in which trees gradually learn structural properties of the data through alternating rounds of generation and discrimination. The method is provably consistent under minimal assumptions. Unlike classic tree-based alternatives, our approach provides smooth (un)conditional densities and allows for fully synthetic data generation. We achieve comparable or superior performance to state-of-the-art probabilistic circuits and deep learning models on various tabular data benchmarks while executing about two orders of magnitude faster on average. An accompanying $\texttt{R}$ package, $\texttt{arf}$, is available on $\texttt{CRAN}$. 
\end{abstract}

\section{INTRODUCTION}

Density estimation is a fundamental unsupervised learning task, an essential subroutine in various methods for data imputation \citep{efron_imputation, rubin1996}, clustering \citep{Bramer2007, Rokach2005}, anomaly detection \citep{chandola2009, pang2021}, and classification \citep{lugosi_1996, pascal2002}. One important application for density estimators is generative modeling, where we aim to create synthetic samples that mimic the characteristics of real data. These simulations can be used to test the robustness of classifiers \citep{song2018, Buzhinsky2021}, augment training sets \citep{ravuri2019, Lopez2018}, or study complex systems without compromising the privacy of data subjects \citep{augenstein_2020, yelmen2021}. 

The current state of the art in generative modeling relies on deep neural networks, which have proven remarkably adept at synthesizing images, audio, and even video data.
Architectures built on variational autoencoders (VAEs) \citep{kingma_vae} and generative adversarial networks (GANs) \citep{goodfellow_gans} have dominated the field for the last decade. Recent advances in normalizing flows \citep{papamakarios2021} and diffusion models \citep{dalle2} have sparked considerable interest.
While these algorithms are highly effective with structured data, they can struggle in tabular settings with continuous and categorical covariates. Even when successful, deep learning models are notoriously data-hungry and require extensive tuning.

Another major drawback of these deep learning methods is that they do not generally permit tractable inference for tasks such as marginalization and conditioning, which are essential for coherent probabilistic reasoning. A family of hierarchical mixture models known as probabilistic circuits (PCs) \citep{vergari2020, choi2020} are better suited to such problems. Despite their attractive theoretical properties, existing PCs can also be slow to train and are often far less expressive than unconstrained neural networks.

We introduce an adversarial random forest algorithm that vastly simplifies the task of density estimation and data synthesis. Our method naturally accommodates mixed data in tabular settings, and performs well on small and large datasets using the computational resources of a standard laptop. It compares favorably with deep learning models while executing some 100 times faster on average. 
It can be compiled into a PC for efficient and exact probabilistic inference.

Following a brief discussion of related work (Sect.~\ref{sec:related}), we review relevant notation and background on random forests (Sect.~\ref{sec:background}). We motivate our method with theoretical results that guarantee convergence under reasonable assumptions (Sect.~\ref{sec:grf}), and illustrate its performance on a range of benchmark tasks (Sect.~\ref{sec:exp}). We conclude with a discussion (Sect.~\ref{sec:discussion}) and directions for future work (Sect.~\ref{sec:conclusion}).

\section{RELATED WORK}\label{sec:related} 

A random forest (RF) is a bootstrap-aggregated (bagged) ensemble of independently randomized trees \citep{Breiman2001}, typically built using the greedy classification and regression tree (CART) algorithm \citep{breiman1984}. 
RFs are extremely popular and effective, widely used in areas like bioinformatics \citep{Chen2012}, remote sensing \citep{belgiu2012}, and ecology \citep{cutler2007}, as well as more generic prediction tasks \citep{fernandez2014}. 
Advantages include their efficiency (RFs are embarrassingly parallelizable), ease of use (they require minimal tuning), and ability to adapt to sparse signals (uninformative features are rarely selected for splits). 

It is well-known that tree-based models can approximate joint distributions. Several authors advocate using leaf nodes of CART trees as piecewise constant density estimators \citep{ram_det2011, wu_rsforest, rfde}. While this method provably converges on the true density in the limit of infinite data, finite sample performance is inevitably rough and discontinuous. Smooth results can be obtained by fitting a distribution within each leaf, e.g. via kernel density estimation (KDE) or maximum likelihood estimation (MLE) \citep{smyth_trees_kde, gray2003, Loh2009, ram_det2011, criminisi2012}, a version of which we develop further below. Existing methods have mostly been limited to supervised trees rather than unsupervised forests, and are often inefficient in high dimensions. 

Another strategy, better suited to high-dimensional settings, uses Chow-Liu trees \citep{chow_liu} to learn a second-order approximation to the underlying joint distribution \citep{bach2003, liu_forest_density, rahman2014}. Whereas these methods estimate a series of bivariate densities over the full support of the data, we attempt to solve a larger number of simpler tasks, modeling univariate densities in relatively small subregions. 

Despite the popularity of tree-based density estimators, they are rarely if ever used for fully synthetic data generation. Instead, they are commonly used for \textit{conditional} density estimation and data imputation \citep{steckhoven2011, Tang2017, Correia2020, Lundberg2020, hothorn2021, cevid2022}. We highlight that methods optimized for this task are often ill-suited to generative modeling, since their reliance on supervised signals limits their ability to capture dependencies between features with little predictive value for the selected outcome variable(s). 

Another family of methods for density estimation and data synthesis is based on probabilistic graphical models (PGMs) \citep{lauritzen1996graphical, koller2009}, e.g. Bayesian networks \citep{pearl2003, darwiche2009}. 
Learning graph structure is difficult in practice, which is why most methods impose restrictive parametric assumptions for tractability \citep{Heckerman1995, drton2017}. 
PCs replace the representational semantics of PGMs with an operational semantics, encoding answers to probabilistic queries in the structural alignment of sum and product nodes. This class of computational graphs subsumes sum-product networks \citep{poon2011}, cutset networks \citep{rahman2014}, and probabilistic sentential decision diagrams \citep{kisa2014}, among others.
\citet{Correia2020} show that RFs instantiate smooth, decomposable, deterministic PCs, thereby enabling efficient marginalization and maximization. 


Deep learning approaches to generative modeling became popular with the introduction of VAEs \citep{kingma_vae} and GANs \citep{goodfellow_gans}, which jointly optimize parameters for network pairs---encoder-decoder and generator-discriminator, respectively---via stochastic gradient descent. Various extensions of these approaches have been developed \citep{higgins_betaVAE, arjovsky_wasserstein_gan}, including some designed for mixed data in the tabular setting \citep{choi_2017, jordon_pategan, Xu2019}. While the evidence lower bound of a VAE approximates the data likelihood, there is no straightforward way to compute this quantity with GANs. More recent work in neural density estimation includes autoregressive networks \citep{pixel_rnn, dalle, roy-ar}, normalizing flows \citep{kobyzev_nf, papamakarios2021, Lee_2022}, and diffusion models \citep{kingma_diffusion, song_diffusion, dalle2}. These methods are generally optimized for structured data such as images or audio, where they often attain state-of-the-art results. 


\section{BACKGROUND}\label{sec:background}

Consider the binary classification setting with training data $\mathcal{D} = \{(\mathbf{x}_i, y_i)\}_{i=1}^n$, where $\mathbf{x}_i \in \mathcal{X} \subset \mathbb{R}^d$ and $y_i \in \mathcal{Y} = \{0, 1\}$. Samples are independent and identically distributed according to some fixed but unknown distribution $P$ with density $p$. The classic RF algorithm takes $B$ bootstrap samples of size $n$ from $\mathcal{D}$ and fits a binary decision tree for each, in which observations are recursively partitioned according to some optimization target (e.g., Gini index) evaluated on a random subset of features at each node. 
The size of this subset is controlled by the $\texttt{mtry}$ parameter, conventionally set at $\lfloor \sqrt{d} \rfloor$ for classification. 
Resulting splits are literals of the form $X_j \bowtie x$ for some $X_j, j \in [d] = \{1, \dots, d\}$, and value $x \in \mathcal{X}_j$, where $\bowtie \; \in \{=,<\}$ (the former for categorical, the latter for continuous variables). Data pass to left or right child nodes depending on whether they satisfy the literal. Splits continue until some stopping criterion is met (e.g., purity). Terminal nodes, a.k.a. \textit{leaves}, describe hyperrectangles in feature space with boundaries given by the learned splits. These disjoint cells collectively cover all of $\mathcal{X}$. 
Each leaf is associated with a label $\hat{y} \in [0, 1]$, representing either the frequency of positive outcomes (soft labels) or the majority class (hard labels) within that cell.
Because trees are grown on independent bootstraps, an average of $n/e$ samples are excluded from each tree. This so-called ``out-of-bag'' (OOB) data can be used to estimate empirical risk without need for cross-validation.

Each new datapoint $\mathbf{x}$ falls into exactly one leaf in each tree. Predictions are computed by aggregating over the trees, e.g. by tallying votes across all $B$ basis functions of the ensemble. Let $\theta^\ell_b$ denote the conjunction of literals that characterize membership in leaf $\ell \in [L_b]$, where $L_b$ is the number of leaves in tree $b \in [B]$, with corresponding hyperrectangular subspace $\mathcal{X}^\ell_b \subset \mathcal{X}$. 
Each leaf has some nonnegative volume and diameter, denoted $\text{vol}(\mathcal{X}^\ell_b)$ and $\text{diam}(\mathcal{X}^\ell_b)$, where the latter represents the longest line segment contained in $\mathcal{X}^\ell_b$.
Let $n_b$ be the number of training samples for tree $b$ (not necessarily equal to $n$) and $n_b^\ell$ the number of samples that fall into leaf $\ell$ of $b$. 
The ratio $n_b^\ell/n_b$ represents an empirical estimate of the leaf's coverage $p(\theta^\ell_b)$, i.e. the probability that a random $\mathbf{x}$ falls within $\mathcal{X}^\ell_b$. 
A tree is fully parametrized by $\bm{\theta}_b = \bigcup_{\ell=1}^{L_b} \theta^\ell_b$, and the complete forest by $\bm{\Theta} = \bigcup_{b=1}^B \bm{\theta}_b$.

Many variations of the classic algorithm exist, including a number of simplified versions designed to be more amenable to statistical analysis. See \citep{Biau2016} for an overview. Common sources of variation include how observations are randomized across trees (e.g., by subsampling or bootstrapping) and how splits are selected (e.g., uniformly or according to some adaptive procedure). 

Our method builds on the \textit{unsupervised} random forest (URF) algorithm \citep{Shi2006}.\footnote{Not to be confused with RF variants that employ non-adaptive splits, which are sometimes also referred to as unsupervised, since they ignore the response variable. See, e.g., \citet{genuer2012}.} This procedure creates a synthetic dataset $\tilde{\mathbf{X}}$ of $n$ observations by independently sampling from the marginals of $\mathbf{X}$, i.e. $\tilde{\mathbf{x}} \sim \prod_{j=1}^d P(X_j)$. A RF classifier is trained to distinguish between $\mathbf{X}$ and $\tilde{\mathbf{X}}$, with labels indicating whether samples are original ($Y=1$) or synthetic ($Y=0$). 
The method has expected accuracy $1/2$ in the worst case, corresponding to a dataset in which all features are mutually independent. However, if dependencies are present, then a consistent learning procedure will converge on expected accuracy $1/2 + \delta$ for some $\delta > 0$ as $n$ grows \citep{kim2021}. 

\section{ADVERSARIAL RANDOM FORESTS}\label{sec:grf}
We introduce a recursive variant of URFs, which we call \textit{adversarial random forest} (ARF). The goal of this algorithm is to render data jointly independent within each leaf. We achieve this by first fitting an ordinary URF $f^{(0)}$ with synthetic data $\tilde{\mathbf{X}}^{(0)}$. We compute the coverage of each leaf w.r.t. original data, then generate a new synthetic dataset by sampling from marginals within random leaves selected with probability proportional to this coverage. Call the resulting $n \times d$ matrix $\tilde{\mathbf{X}}^{(1)}$. A new classifier $f^{(1)}$ is trained to distinguish $\mathbf{X}$ from $\tilde{\mathbf{X}}^{(1)}$. If OOB accuracy for this model is sufficiently low, then the ARF has converged, and we move forward with splits from $f^{(0)}$. Otherwise, we iterate the procedure, drawing a new synthetic dataset from the splits learned by $f^{(1)}$ and evaluating performance via a new classifier. The loop repeats until convergence (see Alg. \ref{alg:arf}).

ARFs bear some obvious resemblance to GANs. The ``generator'' is a simple sampling scheme that draws from the marginals in adaptively selected subregions; the ``discriminator'' is a RF classifier. The result can be understood as a zero-sum game in which adversaries take turns increasing and decreasing label uncertainty at each round. However, beyond this conceptual link between our method and GANs lie some important differences. Both generator and discriminator share the same parameters in our algorithm. Indeed, our generator does not, strictly speaking, \textit{learn} anything; it merely exploits what the discriminator has learned. This means that ARFs cannot be used for adversarial attacks of the sort made famous by GANs, which involve separately parametrized networks for each model. Moreover, the synthetic data generated by ARFs is relatively na\"{i}ve, consisting of bootstrap samples drawn from subsets of the original observations. That is because our goal is not (yet) to generate new data, but merely to learn an independence-inducing partition. Empirically, we find that this is often achieved in just a single round even with the tolerance $\delta$ set to $0$.

\begin{algorithm}[t]
    \small
   \caption{\sc Adversarial Random Forest}
   \label{alg:arf}
\begin{algorithmic}
    \STATE {\bfseries Input:} Training data $\mathbf{X}$, tolerance $\delta$
    \STATE {\bfseries Output:} Random forest classifier $f^{(0)}$
    \STATE 
    \STATE Sample $\tilde{\mathbf{X}}^{(0)} \sim \prod_{j=1}^d P(X_j)$
    \STATE $\mathbf{X}^+ \gets \text{row.append}(\mathbf{X}, \tilde{\mathbf{X}}^{(0)})$
    \STATE $Y \gets \text{row.append}(\mathbf{1}_n, \mathbf{0}_n)$
    \STATE $f^{(0)} \gets \textsc{RandomForest}(\mathbf{X}^+, Y)$
    \IF{$\textsc{Acc}(f^{(0)}) > 1/2 + \delta$}
        \STATE $\texttt{converged} \gets \texttt{FALSE}$
        \WHILE{\textbf{not} \texttt{converged}} 
            \FORALL{$b \in [B^{(0)}], \ell \in [L_b^{(0)}]$}
                \STATE $q(\theta^\ell_b) \gets \frac{2}{n_b} \sum_{i:\mathbf{x}_i \in \mathcal{X}^\ell_b} y_i$
            \ENDFOR
            \FOR{$i \in [n]$} 
                \STATE Sample tree $b \in [B^{(0)}]$ uniformly
                \STATE Sample leaf $\ell \in [L_b^{(0)}]$ w.p. $q(\theta^\ell_b)$
                \STATE Sample $\tilde{\mathbf{x}}^{(1)}_i \sim \prod_{j=1}^d P(X_j | \theta^\ell_b)$
            \ENDFOR
            \STATE $\mathbf{X}^+ \gets \text{row.append}(\mathbf{X}, \tilde{\mathbf{X}}^{(1)})$
            \STATE $f^{(1)} \gets \textsc{RandomForest}(\mathbf{X}^+, Y)$
            \IF{$\textsc{Acc}(f^{(1)}) \leq 1/2 + \delta$}
                \STATE $\texttt{converged} \gets \texttt{TRUE}$
            \ELSE 
                \STATE $f^{(0)} \gets f^{(1)}$
            \ENDIF
        \ENDWHILE
    \ENDIF
\end{algorithmic}
\end{algorithm}

Formally, we seek a set of splits $\bm{\Theta}$ such that, for all trees $b$, leaves $\ell$, and samples $\mathbf{x}$, we have $p(\mathbf{x} | \theta^\ell_b) = \prod_{j=1}^d p(x_j | \theta^\ell_b)$. 
Call this the \textit{local independence criterion}.
Our first result states that ARFs converge on this identity in the limit.
We asssume:
\setlist{leftmargin=10mm}
\begin{itemize}[noitemsep]
    \item[(A1)] The feature domain is limited to $\mathcal{X} = [0,1]^d$, with joint density $p$ bounded away from $0$ and $\infty$. 
    \item[(A2)] At each round, the target function $P(Y=1|\mathbf{x})$ is Lipschitz-continuous. The Lipschitz constant may vary with from one round to the next, but it does not increase faster than $1 / \max_{\ell, b} \big( \text{diam}(\mathcal{X}^\ell_b) \big)$.
    \item[(A3)] Trees satisfy the following conditions: (i) training data for each tree is split into two subsets: one to learn split parameters, the other to assign leaf labels; (ii) trees are grown on subsamples rather than bootstraps, with subsample size $n_b$ satisfying $n_b \rightarrow \infty, n_b / n \rightarrow 0$ as $n \rightarrow \infty$; (iii) at each internal node, the probability that a tree splits on any given $X_j$ is bounded from below by some $\pi > 0$; (iv) every split puts at least a fraction $\gamma \in (0, 0.5]$ of the available observations into each child node; (v) for each tree $b$, the total number of leaves $L_b$ satisfies $L_b \rightarrow \infty, L_b / n \rightarrow 0$ as $n \rightarrow \infty$; and (vi) predictions are made with soft labels both within leaves and across trees, i.e. by averaging rather than voting.
\end{itemize}
(A1) is simply for notational convenience, and can be replaced w.l.o.g. by bounding the feature domain with arbitrary constants.
Lipschitz continuity is a common learning theoretic assumption widely used in the analysis of RFs. 
(A2)'s extra condition regarding the Lipschitz constant controls the variation in smoothness over adversarial training rounds.
(A3) imposes standard regularity conditions for RFs \citep{Meinshausen2006, biau_analysis, denil2014, scornet2016_asymptotics, Wager2018}.  
With these assumptions in place, we have the following result (see Appx.~\ref{sec:app_proof} for all proofs).

\begin{theorem}[Convergence] \label{thm:conv}
    Under (A1)-(A3), ARFs converge in probability on the local independence criterion. 
    Let $\Theta_n$ be the parameters of an ARF trained on a sample of size $n$. 
    Then for all $\mathbf{x} \in \mathcal{X}$, $\theta^\ell_b \in \Theta_n$, and $\epsilon > 0$:
    \begin{align*}
        \lim_{n \rightarrow \infty} \mathbb{P} \Big[ \big| p(\mathbf{x} | \theta^\ell_b) - \prod_{j=1}^d p(x_j | \theta^\ell_b) \big| \geq \epsilon \Big] = 0.
    \end{align*}
\end{theorem}


\begin{algorithm}[t]
    \small
   \caption{\sc Forde}
   \label{alg:ford}
\begin{algorithmic}
    \STATE {\bfseries Input:} ARF classifier $f$, training data $\mathbf{X} \in \mathbb{R}^{n \times d}$
    \STATE {\bfseries Output:} Estimated density $q$
    \STATE
    \FORALL{$b \in [B], \ell \in [L_b]$}
        \STATE $q(\theta^\ell_b) \gets \frac{2}{n_b} \sum_{i:\mathbf{x}_i \in \mathcal{X}^\ell_b} y_i$
        \FOR{$j \in [d]$}
            \STATE $\bm{\psi}_{b,j}^\ell \gets \text{estimated parameter(s) for}~ p(x_j|\theta^\ell_b)$
            \STATE $q(\cdot~; \bm{\psi}_{b,j}^\ell) \gets \text{corresponding pdf/pmf}$
        \ENDFOR 
    \ENDFOR
\end{algorithmic}
\end{algorithm}

\begin{algorithm}[t]
    \small
   \caption{\sc Forge}
   \label{alg:forge}
\begin{algorithmic}
    \STATE {\bfseries Input:} $\textsc{Forde}$ model $q$, target sample size $m$
    \STATE {\bfseries Output:} Synthetic dataset $\tilde{\mathbf{X}} \in \mathbb{R}^{m \times d}$
    \STATE
    \FOR{$i \in [m]$}
        \STATE $\text{Sample tree}~ b \in [B] ~\text{uniformly}$
        \STATE $\text{Sample leaf}~ \ell \in [L_b] ~\text{w.p.}~ q(\theta^\ell_b)$
        \FOR{$j \in [d]$}
            \STATE $\text{Sample data}~ \tilde{x}_{ij} \sim q(\cdot~; \bm{\psi}_{b,j}^\ell)$
        \ENDFOR
    \ENDFOR
\end{algorithmic}
\end{algorithm}

\subsection{Density Estimation and Data Synthesis}
ARFs are the basis for two further algorithms, FORests for Density Estimation ($\textsc{Forde}$) and FORests for GEnerative modeling ($\textsc{Forge}$). We present pseudocode for both (see Algs. \ref{alg:ford} and \ref{alg:forge}). The key point to recognize is that, under the local independence criterion, joint densities can be learned by running $d$ separate univariate estimators within each leaf. This is exponentially easier than multivariate density estimation, which suffers from the notorious \textit{curse of dimensionality}. Summarizing the challenges with estimating joint densities, one recent textbook on KDE concludes that ``nonparametric methods for kernel density problems should not be used for high-dimensional data and it seems that a feasible dimensionality should not exceed five or six...'' \citep[p. 60]{Gramacki_kde}. By contrast, our method scales much better with data dimensionality, exploiting the flexibility of ARFs to learn an independence-inducing partition that renders density estimation relatively straightforward. 

Of course, this does not \textit{escape} the curse of dimensionality so much as relocate it. The cost for this move is potentially deep trees and/or many ARF training rounds, especially when dependencies between covariates are strong or complex. However, deep forests are generally more efficient than deep neural networks in terms of data and computation, and our experiments suggest that ARF convergence is usually fast even for $\delta = 0$ (see Sect.~\ref{sec:exp}).

With our ARF in hand, the algorithm proceeds as follows. For each tree $b$, we record the split criteria $\theta^\ell_b$ and empirical coverage $q(\theta^\ell_b)$ of each leaf $\ell$. Call these the \textit{leaf parameters}. Then we estimate \textit{distribution parameters} $\bm{\psi}_{b,j}^\ell$ independently for each (original) $X_j$ within $\mathcal{X}^\ell_b$, e.g. the kernel bandwidth for KDE or class probabilities for MLE with categorical data. 
In the continuous case, $\bm{\psi}_{b,j}^\ell$ must either encode leaf bounds (e.g., via a truncated normal distribution with extrema given by $\theta^\ell_b$) or include a normalization constant to ensure integration to unity.
The generative model then follows a simple two-step procedure. First, sample a tree uniformly from $[B]$ and a leaf from that tree with probability $q(\theta^\ell_b)$, just as we do to construct synthetic data within the recursive loop of the ARF algorithm. Next, sample data for each feature $X_j$ according to the density/mass function parametrized by $\bm{\psi}_{b,j}^\ell$. We repeat this procedure until the target number of synthetic samples has been generated.




We are deliberately agnostic about how distribution parameters $\bm{\psi}_{b,j}^\ell$ should be learned, as this will tend to vary across features. In our theoretical analysis, we restrict focus to continuous variables and consider a flexible family of KDE methods. In our experiments, we use MLE for continuous data, effectively implementing a truncated Gaussian mixture model, and Bayesian inference for categorical variables, to avoid extreme probabilities when values are unobserved but not beyond the support of a given leaf.
Under local independence, distribution learning is completely modular, so different methods can coexist without issue. We revisit this topic in Sect.~\ref{sec:discussion}.

Our estimated density takes the following form:
\begin{equation}\label{eq:est}
    q(\mathbf{x}) = \frac{1}{B} \sum_{\ell, b: \mathbf{x} \in \mathcal{X}^\ell_b} ~q(\theta^\ell_b) ~\prod_{j=1}^d q(x_{j}; \bm{\psi}_{b,j}^\ell).
\end{equation}
Compare this with the true density:
\begin{equation}\label{eq:tru}
    p(\mathbf{x}) = \frac{1}{B} \sum_{\ell, b: \mathbf{x} \in \mathcal{X}^\ell_b} p(\theta^\ell_b) ~p(\mathbf{x}| \theta^\ell_b).
\end{equation}
In both cases, the density evaluated at a given point is just a coverage-weighted average of its density in all leaves whose split criteria it satisfies. 

Because we are concerned with $L_2$-consistency, our loss function is the mean integrated squared error (MISE)\footnote{Alternative loss functions may also be suitable, e.g. the Kullback-Leibler divergence or the Wasserstein distance.}, defined as: 
\begin{equation*}\label{eq:ise}
    \text{MISE}(p, q) := \mathbb{E} \Bigg[\int_{\mathcal{X}} \Big( p(\mathbf{x}) - q(\mathbf{x}) \Big)^2 ~d\mathbf{x} \Bigg].
\end{equation*}

We require one extra assumption, imposing standard conditions for KDE consistency \citep{silverman1986}:
\begin{itemize}[noitemsep]
    \item[(A4)] The true density function $p$ is smooth. Specifically, its second derivative $p''$ is finite, continuous, square integrable, and ultimately monotone.
\end{itemize}
Our method admits three potential sources of error, quantified by the following residuals: 
\begin{align}
    \epsilon_1 &:= \epsilon_1(\ell, b) :=  p(\theta^\ell_b) - q(\theta^\ell_b) \label{eq:eps1}\\
    \epsilon_2 &:= \epsilon_2(\ell, b, \mathbf{x}) := \prod_{j=1}^d p(x_j|\theta^\ell_b) - \prod_{j=1}^d q(x_j; \bm{\psi}_{b,j}^{\ell}) \label{eq:eps2}\\
    \epsilon_3 &:= \epsilon_3(\ell, b, \mathbf{x}) := p(\mathbf{x} | \theta^\ell_b) - \prod_{j=1}^d p(x_j|\theta^\ell_b) \label{eq:eps3}
\end{align}
We refer to these as errors of \textit{coverage}, \textit{density}, and \textit{convergence}, respectively. Observe that $\epsilon_1$ is a random variable that depends on $\ell$ and $b$, while $\epsilon_2, \epsilon_3$ are random variables depending on $\ell, b$ and $\mathbf{x}$. We suppress the dependencies for ease of notation. 
\begin{lemma}\label{lm:decomp}
The error of our estimator satisfies the following bound:
\begin{align*}
    \text{MISE}(p, q) \leq 2B^{-2}~\mathbb{E} \Bigg[ \int_\mathcal{X} \alpha^2 + \beta^2 ~d\mathbf{x} \Bigg],
\end{align*}
where
\begin{align*}
    \alpha &:= \sum_{\ell, b: \mathbf{x} \in \mathcal{X}_b^\ell} p(\theta^\ell_b) \epsilon_3 \quad \text{and} \\
    \beta &:= \sum_{\ell, b: \mathbf{x} \in \mathcal{X}_b^\ell} \Big( p(\theta^\ell_b)\epsilon_2 + \epsilon_1\prod_{j=1}^d p(x_j|\theta^\ell_b) - \epsilon_1 \epsilon_2 \Big).
\end{align*}
\end{lemma}
This lemma establishes that total error is bounded by a quadratic function of $\epsilon_1, \epsilon_2, \epsilon_3$. We know by Thm. \ref{thm:conv} that errors of convergence vanish in the limit. Our next result states that the same holds for errors of coverage and density.

\begin{theorem}[Consistency] \label{thm:cons}
    Under assumptions (A1)-(A4), $\textsc{Forde}$ is $L_2$-consistent. Let $q_n$ denote the joint density estimated on a training sample of size $n$. Then we have:
    \begin{align*}
        \lim_{n \rightarrow \infty} \normalfont{\text{MISE}}(p, q_n) = 0.
    \end{align*}
\end{theorem}
Our consistency proof is fundamentally unlike those of piecewise constant density estimators with CART trees \citep{ram_det2011, wu_rsforest, Correia2020}, which essentially treat base learners as adaptive histograms and rely on tree-wise convergence when leaf volume goes to zero \citep{devroye1996, lugosi_1996}. Alternative methods that perform KDE or MLE within each leaf do not come with theoretical guarantees \citep{smyth_trees_kde, gray2003, Loh2009, ram_det2011}. 
Recently, consistency has been shown for some RF-based conditional density estimators \citep{hothorn2021, cevid2022}. However, these results do not extend to the unconditional case, since features with little predictive value for the outcome variable(s) are unlikely to be selected for splits. The resulting models will therefore fail to detect dependencies between features deemed uninformative for the given prediction task.  
In the rare case that authors use some form of unsupervised splits, they make no effort to factorize the distribution and are therefore subject to the curse of dimensionality \citep{criminisi2012, feng_autoencoder}. By contrast, our method exploits ARFs to find regions of local independence, and univariate density estimation to compute marginals within each leaf. Though our consistency result comes at the cost of some extra assumptions, we argue that this is a fair price to pay for improved performance in finite samples.



\begin{figure}[t]
  \centering
  \includegraphics[width=0.45\textwidth]{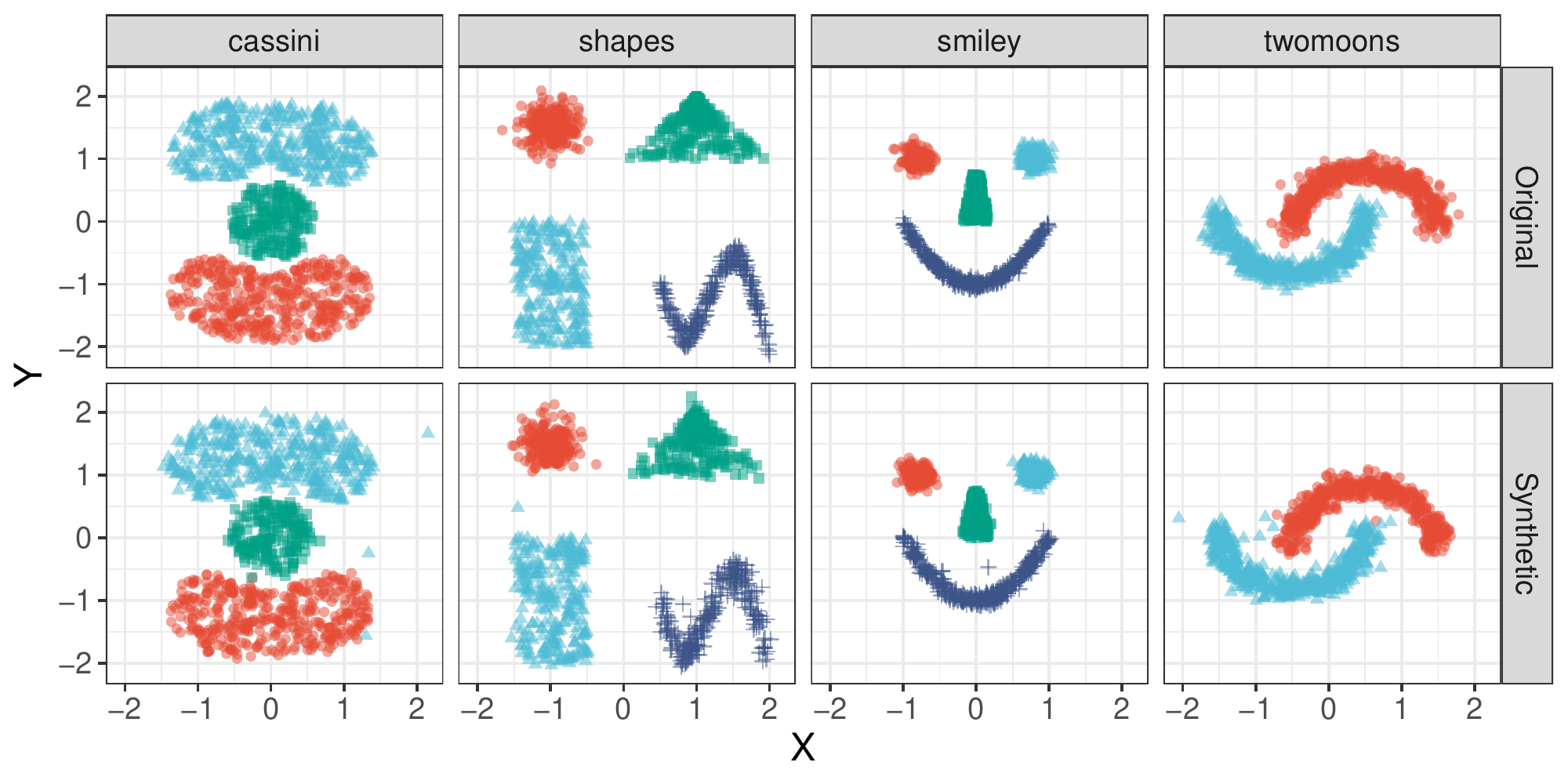}
  \vspace{-2mm}
  \caption{Visual examples. Original (top) and synthetic (bottom) data are presented for four three-dimensional problems with two continuous covariates and one categorical feature.}
  \label{fig:examples}
\end{figure}

\section{EXPERIMENTS}\label{sec:exp}

In this section, we present results from a wide range of experiments conducted on simulated and real-world datasets. We use 100 trees for density estimation tasks and 20 for data synthesis. Increasing this parameter tends to improve performance for $\textsc{Forde}$, but appears to have less of an impact on $\textsc{Forge}$. 
Trees are grown until purity or a minimum node size of two (with just a single sample, variance is undefined). 
In all cases, we set the slack parameter $\delta = 0$ and use the default $\texttt{mtry} = \lfloor \sqrt{d} \rfloor$.
For more details on hyperparameters and datasets see Appx. \ref{sec:app_exp}. 
Code for reproducing all results is available online at \url{https://github.com/bips-hb/arf_paper}.

\subsection{Simulation}

\paragraph{$\textsc{Forge}$ recreates visual patterns.}
We begin with a simple proof of concept experiment, illustrating our method on a handful of low-dimensional datasets that allow for easy visual assessment. The $\texttt{cassini}$, $\texttt{shapes}$, $\texttt{smiley}$, and $\texttt{twomoons}$ problems are all three-dimensional examples that combine two continuous covariates with a categorical class label. We simulate $n=2000$ samples from each data generating process (see Fig.~\ref{fig:examples}, top row) and estimate densities using $\textsc{Forde}$. We proceed to $\textsc{Forge}$ a synthetic dataset of $n=1000$ samples (Fig.~\ref{fig:examples}, bottom row) and compare results. We find that the model consistently approximates its target distribution with high fidelity. Classes are clearly distinguished in all cases, and the visual form of the original data is immediately recognizable. A few stray samples are evident on close inspection. Such anomalies can be mitigated with a larger training set.  

\paragraph{$\textsc{Forde}$ outperforms alternative CART-based methods.}
We simulate data from a multivariate Gaussian distribution $\mathbf{X} \sim \mathcal{N}(0,\Sigma)$, with Toeplitz covariance matrix $\Sigma_{ij} = 0.9^{|i-j|}$ and fixed $d=10$. To compare against supervised methods, we also simulate a binary target $Y \sim \text{Bern}([1 + \exp(-\textbf{X} \bm{\beta})]^{-1})$, where the coefficient vector $\bm{\beta}$ contains a varying proportion of $0$'s (non-informative features) and $1$'s (informative features). Performance is evaluated by the negative log-likelihood (NLL) on a test set of $n_{\text{tst}}=1000$. 
We compare our method to piecewise constant (PWC) estimators with supervised and unsupervised split criteria,\footnote{Supervised PWC is simply an ensemble version of the classic method \citep{gray2003, ram_det2011, wu_rsforest}. To the best of our knowledge, no one has previously proposed \textit{unsupervised} PWC density estimation with CART trees. This can be understood as a variant of our approach in which all marginals are uniform within each leaf.} as well as generative forests (GeFs), a RF-based smooth density estimation procedure \citep{Correia2020}. 


\begin{figure}[t]
  \centering
  \includegraphics[width=0.48\textwidth]{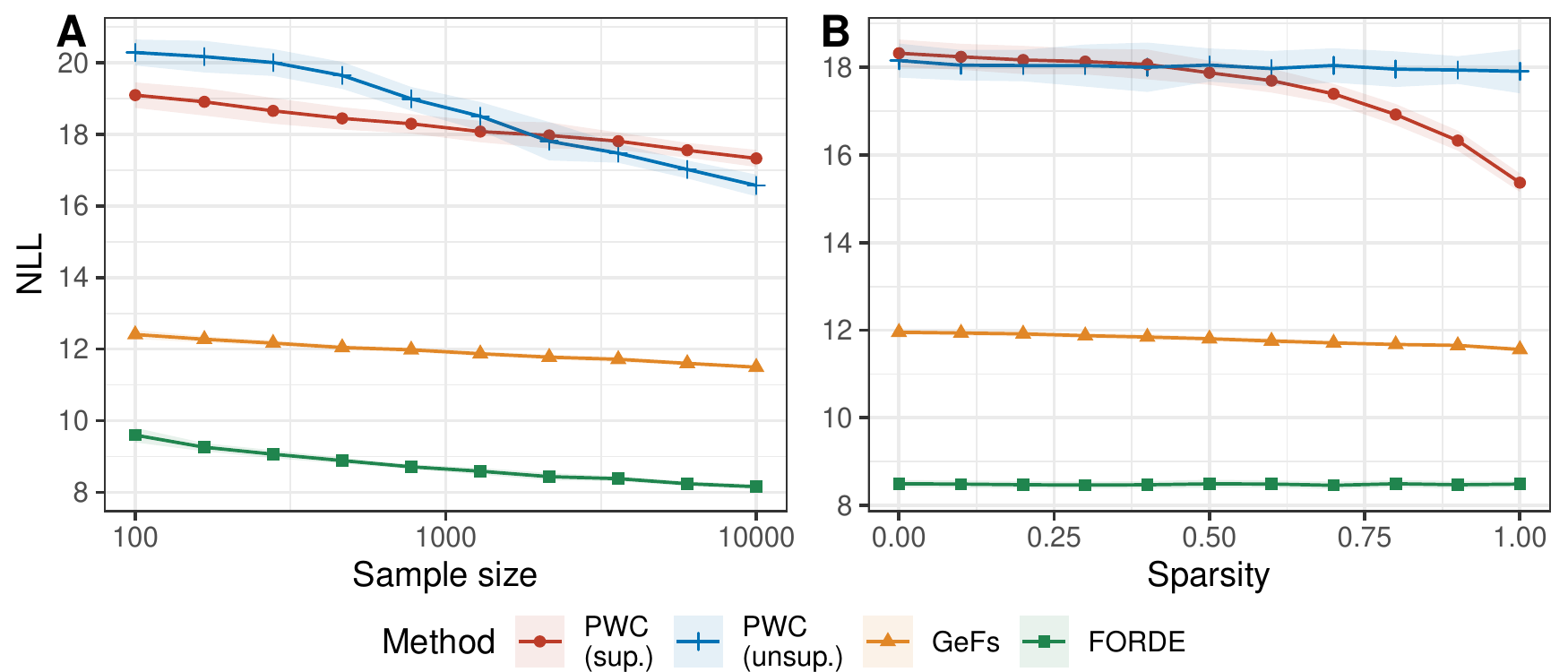}
  \vspace{-5mm}
  \caption{Negative log-likelihood (NLL) measured in nats on a test set for varying sample size \textbf{(A)} and sparsity \textbf{(B)}. Lower is better. Shading represents standard errors.}
  \label{fig:nll}
\end{figure}

Fig.~\ref{fig:nll} shows the average NLL over $20$ replicates for varying sample sizes (A) and levels of sparsity (B). For the former, we fix the proportion of informative features at 0.5; for the latter, we fix $n_{\text{trn}}=2000$. We find that PWC methods fare poorly, with much greater NLL in all settings. This is likely due to the unrealistic uniformity assumption, according to which the corner of a hyperrectangle is no less probable than the center. GeFs, which also use a Gaussian mixture model to estimate densities, perform better in this experiment. However, $\textsc{Forde}$ dominates throughout. 

Panel (B) clearly illustrates that unsupervised methods are unaffected by changes in signal sparsity, since their splits are independent of the outcome variable $Y$. By contrast, sparsity appears to benefit the supervised methods. This can be explained by the fact that splits are random when features are uninformative, a strategy that is known to work well in noisy settings \citep{Geurts2006, genuer2012}.


\begin{table}[t]
\tiny
\centering
\caption{Average NLL on the Twenty Datasets benchmark for five PC models and $\textsc{Forde}$. Winning results in bold.}
\label{tbl:20d}
\begin{tabular}{lrrrrrr}
  \toprule
Dataset   & EiNet & RAT-SPN  & PGC    & Strudel   & CMCLT  & \textsc{Forde}  \\
  \midrule
\texttt{nltcs}     & 6.02    & 6.01   & 6.05   & 6.07   & \textbf{5.99}   & 6.01\\
\texttt{msnbc}     & 6.12    & \textbf{6.04}   & 6.06   & \textbf{6.04}   & 6.05   & 6.10\\
\texttt{kdd}       & 2.18    & 2.13   & 2.14   & 2.14   & \textbf{2.12}   & 2.13\\
\texttt{plants}    & 13.68   & 13.44  & 13.52  & 13.22  & \textbf{12.26}  & \textbf{12.26}\\
\texttt{audio}     & 39.88   & 39.96  & 40.21  & 42.40  & \textbf{39.02}  & 39.74\\
\texttt{jester}    & 52.56   & 52.97  & 53.54  & 54.24  & \textbf{51.94}  & 52.8\\
\texttt{netflix}   & 56.64   & 56.85  & 57.42  & 57.93  & \textbf{55.31}  & 56.67\\
\texttt{accidents} & 35.59   & 35.49  & 30.46  & 29.05  & \textbf{28.69}  & 33.85\\
\texttt{retail}    & 10.92   & 10.91  & 10.84  & 10.83  & \textbf{10.82}  & 10.93\\
\texttt{pumsb}     & 31.95   & 31.53  & 29.56  & 24.39  & \textbf{23.71}  & 28.4\\
\texttt{dna}       & 96.09   & 97.23  & \textbf{80.82}  & 87.15  & 84.91  & 91.85\\
\texttt{kosarek}   & 11.03   & 10.89  & 10.72  & 10.70  & \textbf{10.56}  & 10.84\\
\texttt{msweb}     & 10.03    & 10.12  & 9.98   & 9.74   & \textbf{9.62}   & 9.72\\
\texttt{book}      & 34.74   & 34.68  & 34.11  & 34.49  & \textbf{33.75}  & 34.85\\
\texttt{movie}     & 51.71   & 53.63  & 53.15  & 53.72  & \textbf{49.23}  & 50.86\\
\texttt{webkb}     & 157.28  & 157.53 & 155.23 & 154.83 & \textbf{147.77} & 153.45\\
\texttt{reuters}   & 87.37   & 87.37  & 87.65  & 86.35  & \textbf{81.17}  & 84.15\\
\texttt{20ng}      & 153.94  & 152.06 & 154.03 & 153.87 & \textbf{148.17} & 155.51\\
\texttt{bbc}       & 248.33  & 252.14 & 254.81 & 256.53 & 242.83 & \textbf{240.31}\\
\texttt{ad}        & 26.27   & 48.47  & 21.65  & 16.52  & \textbf{14.76}  & 21.80\\
\midrule
Avg. rank & 4.5 & 4.2 & 4 & 3.6 & 1.2 & 3.3 \\
  \bottomrule
\end{tabular}
\end{table}

\subsection{Real Data}\label{sec:benchmark}

\textbf{$\textsc{Forde}$ is competitive with alternative PCs.}
Building on \citet{Correia2020}'s observation that RFs can be compiled into probabilistic circuits, we compare the performance of $\textsc{Forde}$ to that of five leading PCs on the Twenty Datasets benchmark \citep{Davis_2021}, a heterogeneous collection of tasks ranging from retail to biology that is widely used to evaluate tractable probabilistic models. 
Each dataset is randomly split into training (70\%), validation (10\%), and test sets (20\%). Competitors include Einsum networks (EiNet) \citep{peharz2020}, random sum-product networks (RAT-SPN) \citep{ratspn}, probabilistic generating circuits (PGC) \citep{zhang2021}, Strudel \citep{dang2022}, and continuous mixtures of Chow-Liu trees (CMCLT) \citep{correia2022}. We report the average NLL on the test set for each model in Table \ref{tbl:20d}. Though the recently proposed CMCLT algorithm generally dominates in this experiment, $\textsc{Forde}$ attains top performance on two datasets and is never far behind the state of the art. Its average rank of 3.3 places it second overall.


\textbf{$\textsc{Forge}$ generates realistic tabular data.}
To evaluate the performance of $\textsc{Forge}$ on real-world datasets, we recreate a benchmarking pipeline originally proposed by \citet{Xu2019}. They introduce the conditional tabular GAN ($\textsc{ctgan}$) and tabular VAE ($\textsc{tvae}$), two deep learning algorithms for generative modeling with mixed continuous and categorical features. We include three additional state-of-the-art tabular GAN architectures for comparison: invertible tabular GAN ($\textsc{it-gan}$) \citep{lee2021_it-gan}, regularized compound conditional GAN ($\textsc{rcc-gan}$) \citep{Esmaeilpour2022}, and a differentially private conditional tabular GAN ($\textsc{ctab-gan+}$) \citep{zhao2022}.

A complete summary of the experimental setup is presented in Appx.~\ref{sec:app_exp}. Briefly, we take five benchmark datasets for classification and partition the samples into training and test sets, which we denote by $\mathbf{Z}_{\text{trn}} = (\mathbf{X}_{\text{trn}}, Y_{\text{trn}})$ and $\mathbf{Z}_{\text{tst}} = (\mathbf{X}_{\text{tst}}, Y_{\text{tst}})$, respectively. 
$\mathbf{Z}_{\text{trn}}$ is used as input to a series of generative models, each of which creates a synthetic training set $\tilde{\mathbf{Z}}_{\text{trn}}$ of the same sample size as the original. Several classifiers are then trained on $\tilde{\mathbf{Z}}_{\text{trn}}$ and evaluated on $\mathbf{Z}_{\text{tst}}$, with performance metrics averaged across learners. Results are benchmarked against the same set of algorithms, now trained on the original data $\mathbf{Z}_{\text{trn}}$. We refer to this model as the \textit{oracle}, since it should perform no worse in expectation than any classifier trained on synthetic data. However, if the generative model approximates its target with high fidelity, then differences between the oracle and its competitors should be negligible.\footnote{Note that the so-called ``oracle'' is not necessarily optimal w.r.t. the true data generating process---other models may have lower risk---but it should be optimal w.r.t. a given function class-dataset pair. If logistic regression attains 60\% test accuracy training on $\mathbf{Z}_{\text{trn}}$, then it should do about the same training on $\tilde{\mathbf{Z}}_{\text{trn}}$, regardless of how much better a well-tuned MLP may perform.} Similar approaches are widely used in the evaluation of GANs \citep{lr_gan, how_good_gan, santurkar2018}; for a critical discussion, see \cite{ravuri2019}.

Results are reported in Table \ref{benchmark}, where we average over five trials of data synthesis and subsequent supervised learning. We include information on each dataset, including the cardinality of the response variable, the training/test sample size, and dimensionality of the feature space. Performance is evaluated via accuracy and F1-score (or F1 macro-score for multiclass problems), as well as wall time. 
$\textsc{Forge}$ fares well in this experiment, attaining the top accuracy and F1-score in three out of five tasks. 
On a fourth, the highly imbalanced $\texttt{credit}$ dataset, the only models that do better in terms of accuracy receive F1-scores of 0, suggesting that they entirely ignore the minority class. Only $\textsc{Forge}$ and $\textsc{rcc-gan}$ strike a reasonable balance between sensitivity and specificity on this task.
Perhaps most impressive, $\textsc{Forge}$ executes over 60 times faster than its nearest competitor on average, and over 100 times faster than the second fastest method. (We omit results for algorithms that fail to converge in 24 hours of training time.)
Differences in compute time would be even more dramatic if these deep learning algorithms were configured with a CPU backend (we used GPUs here), or if $\textsc{Forge}$ were run using more extensive parallelization (we distribute the job across 10 cores). This comparison also obscures the extra time required to tune hyperparameters for these complex models, whereas our method is an off-the-shelf solution that works well with default settings. 

\begin{table}[t]
\tiny
\centering
\caption{Performance on the \citet{Xu2019} benchmark for five deep learning models and $\textsc{Forge}$. We report average results across five replicates $\pm$ standard errors. Winning results in bold.  
} 
\label{benchmark}
\begin{tabular}{llccr}
  \toprule
Dataset & Model & Accuracy $\pm$ SE  & F1 $\pm$ SE & Time (sec) \\ 
  \midrule
\texttt{adult} & \textit{Oracle} & \textit{0.828 $\pm$ 0.006} & \textit{0.884 $\pm$ 0.004} & \\ 
   classes = 2 & $\textsc{Forge}$ & \textbf{0.819} $\pm$ 0.006 & \textbf{0.877} $\pm$ 0.005 & \textbf{2.9} \\ 
   $n_{\text{trn}} $ = 23k & $\textsc{ctgan}$ & 0.786 $\pm$ 0.020 & 0.853 $\pm$ 0.019 & 263.3 \\ 
   $n_{\text{tst}} $ = 10k & $\textsc{ctab-gan+}$ & 0.808 $\pm$ 0.008 & 0.869 $\pm$ 0.006 & 561.6 \\ 
   $d$ = 14& $\textsc{it-gan}$ & 0.794 $\pm$ 0.005 & 0.853 $\pm$ 0.005 & 3435.6 \\ 
   & $\textsc{rcc-gan}$ & 0.770 $\pm$ 0.015 & 0.841 $\pm$ 0.015 & 8823.0 \\ 
   & $\textsc{tvae}$ & 0.804 $\pm$ 0.007 & 0.865 $\pm$ 0.006 & 115.1 \\ \midrule
  \texttt{census} & \textit{Oracle} & \textit{0.922 $\pm$ 0.002} & \textit{0.957 $\pm$ 0.001} &  \\ 
   classes = 2 & $\textsc{Forge}$ & 0.903 $\pm$ 0.019 & 0.946 $\pm$ 0.012 & \textbf{53.2} \\ 
   $n_{\text{trn}} $ = 200k & $\textsc{ctgan}$ & 0.916 $\pm$ 0.015 & 0.954 $\pm$ 0.009 & 4287.8\\ 
   $n_{\text{tst}} $ = 100k & $\textsc{ctab-gan+}$ & 0.912 $\pm$ 0.026 & 0.952 $\pm$ 0.016 & 10182.1 \\ 
   $d$ = 40& $\textsc{it-gan}$ & NA   & NA    & $>$24hr \\ 
   & $\textsc{rcc-gan}$ & 0.900 $\pm$ 0.016 & 0.944 $\pm$ 0.011 & 8908.6 \\ 
   & $\textsc{tvae}$ & \textbf{0.928} $\pm$ 0.007 & \textbf{0.961} $\pm$ 0.004 & 1814.9 \\ \midrule
  \texttt{covertype} & \textit{Oracle} & \textit{0.895 $\pm$ 0.000} & \textit{0.838 $\pm$ 0.000} & \\ 
   classes = 7 & $\textsc{Forge}$ & \textbf{0.707} $\pm$ 0.006 & \textbf{0.549} $\pm$ 0.006 & \textbf{103.5} \\ 
   $n_{\text{trn}}$ = 481k & $\textsc{ctgan}$ & 0.633 $\pm$ 0.009 & 0.400 $\pm$ 0.009 & 13387.2 \\ 
   $n_{\text{tst}} $ = 100k & $\textsc{ctab-gan+}$ & NA   & NA    & $>$24hr \\ 
   $d$ = 54 & $\textsc{it-gan}$ & NA   & NA    & $>$24hr \\ 
   & $\textsc{rcc-gan}$ & NA   & NA    & $>$24hr \\ 
   & $\textsc{tvae}$ & 0.698 $\pm$ 0.013 & 0.459 $\pm$ 0.013 & 4882.0 \\ \midrule
  \texttt{credit} & \textit{Oracle} & \textit{0.997 $\pm$ 0.001} & \textit{0.607 $\pm$ 0.029} &  \\ 
   classes = 2 & $\textsc{Forge}$ & 0.995 $\pm$ 0.001 & 0.527 $\pm$ 0.036 & \textbf{32.2} \\ 
   $n_{\text{trn}}$ = 264k & $\textsc{ctgan}$ & 0.881 $\pm$ 0.099 & 0.047 $\pm$ 0.031 & 4898.0 \\ 
   $n_{\text{tst}}$ = 20k & $\textsc{ctab-gan+}$ & \textbf{0.998} $\pm$ 0.000 & 0.000 $\pm$ 0.000 & 7497.3 \\ 
   $d$ = 30& $\textsc{it-gan}$ & NA   & NA    & $>$24hr \\ 
   & $\textsc{rcc-gan}$ & 0.993 $\pm$ 0.003 & \textbf{0.569} $\pm$ 0.056 & 10608.4 \\ 
   & $\textsc{tvae}$ & \textbf{0.998} $\pm$ 0.000 & 0.000 $\pm$ 0.000 & 3847.6 \\ \midrule
  \texttt{intrusion} & \textit{Oracle} & \textit{0.998 $\pm$ 0.001} & \textit{0.833 $\pm$ 0.001} &  \\ 
   classes = 5 & $\textsc{Forge}$ & \textbf{0.993} $\pm$ 0.001 & \textbf{0.656} $\pm$ 0.001 & \textbf{68.2} \\ 
   $n_{\text{trn}}$ = 394k & $\textsc{ctgan}$ & 0.944 $\pm$ 0.088 & 0.645 $\pm$ 0.088 & 8749.3 \\ 
   $n_{\text{tst}} $ = 100k& $\textsc{ctab-gan+}$ & NA   & NA    & $>$24hr \\ 
   $d$ = 40 & $\textsc{it-gan}$ & NA   & NA    & $>$24hr \\ 
   & $\textsc{rcc-gan}$ & NA   & NA    & $>$24hr \\
   & $\textsc{tvae}$ & 0.990 $\pm$ 0.002 & 0.598 $\pm$ 0.002 & 4306.0 \\ 
   \bottomrule
\end{tabular}
\end{table}

\subsection{Runtime}\label{sec:runtime}

To further demonstrate the computational efficiency of our pipeline relative to deep learning methods, we conduct a runtime experiment using the smallest dataset above, $\texttt{adult}$. By repeatedly sampling stratified subsets---varying both sample size $n$ and dimensionality $d$---and measuring the time needed to train a generative model and synthesize data from it, we illustrate how complexity scales with $n$ and $d$. 
For this experiment, we ran the three fastest deep learning competitors---$\textsc{ctgan}$, $\textsc{tvae}$, and $\textsc{ctab-gan+}$---with both CPU and GPU backends. We use default parameters for all algorithms, which include automated parallelization over all available cores (24 in this experiment). 

\begin{figure*}[t]
  \centering
  \includegraphics[width=0.85\textwidth]{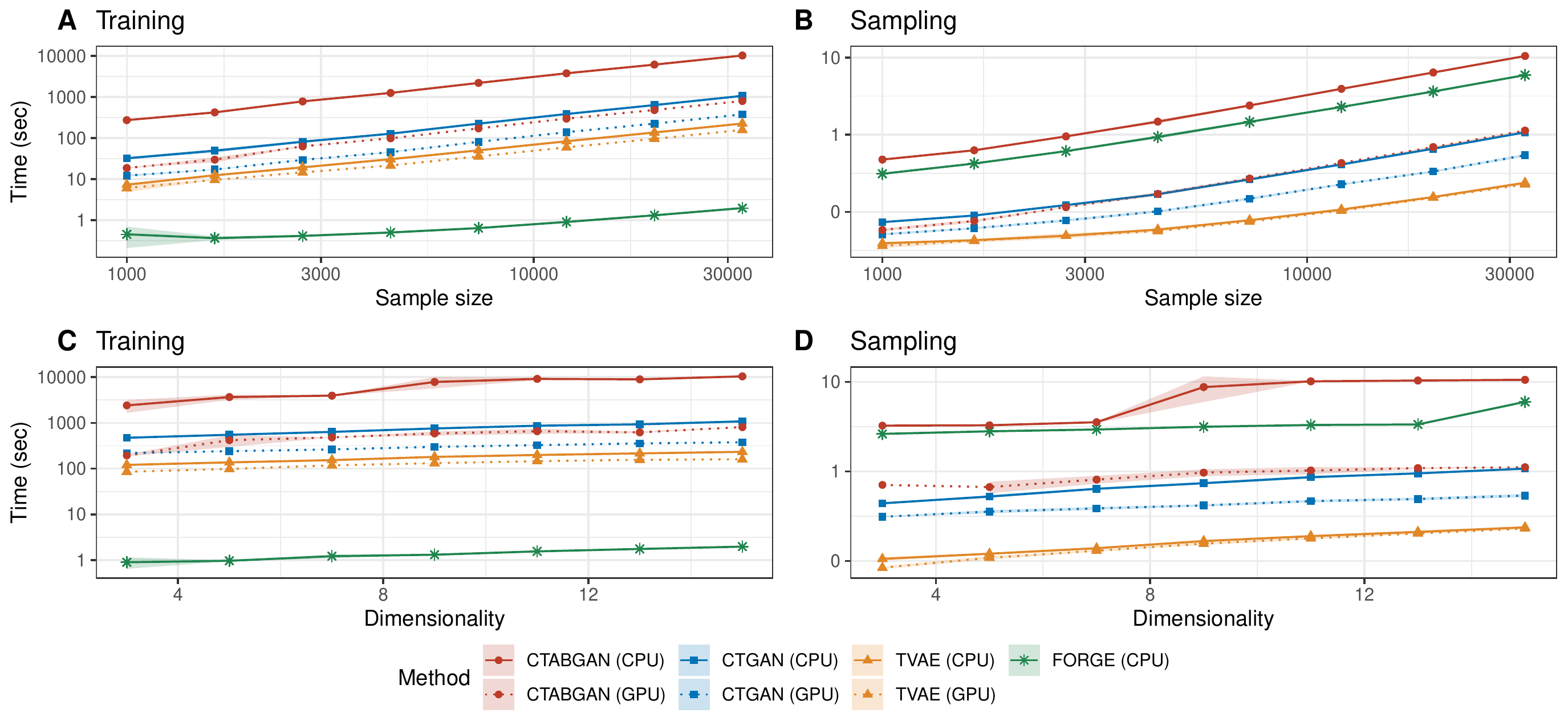}
  \caption{Complexity curves, evaluated using stratified subsamples of the \texttt{adult} dataset. \textbf{(A)}: Training time as a function of sample size.  \textbf{(B)}: Sampling time as a function of sample size. \textbf{(C)}: Training time as a function of dimensionality. \textbf{(D)}: Sampling time as a function of dimensionality.}
  \label{fig:time}
  \vspace{-5mm}
\end{figure*}

Fig.~\ref{fig:time} shows the results. $\textsc{Forge}$ clearly dominates in training time (see panels A and C), executing orders of magnitude faster than the competition (note the log scale). For those with limited access to GPUs, deep learning methods may be completely infeasible for large datasets. Even when GPUs are available, $\textsc{Forge}$ still scales far better, completing the full pipeline about 35 times faster than $\textsc{tvae}$, 85 times faster than $\textsc{ctgan}$, and nearly 200 times faster than $\textsc{ctab-gan+}$ in this example. 
Other methods appear to generate samples more quickly than $\textsc{Forge}$ (see panels B and D), but this computation is trivial compared to training. Interestingly, our method is a faster sampler when measured in processing time (see Fig. \ref{fig:process_time}, Appx. \ref{app:runtime}), suggesting that it could outperform competitors here too with more efficient parallelization.
Note that $\textsc{Forge}$ attains the highest accuracy and F1-score of all methods for the $\texttt{adult}$ dataset, so this speedup need not come at the cost of performance. 

\section{DISCUSSION}\label{sec:discussion}


ARFs enable fast, accurate density estimation and data synthesis. However, the method is not without its limitations. First, it is not tailored to structured data such as images or text, for which deep learning models have proven especially effective. See Appx.~\ref{sec:mnist} for a comparison to state-of-the-art models on the MNIST dataset, where convolutional GANs clearly outperform $\textsc{Forge}$, as expected. Where our method excels, by contrast, is in speed and flexibility.

We caution that our convergence guarantees have no implications for finite sample performance. Though ARFs only required a few rounds of training in most of our experiments, it is entirely possible that discriminator accuracy increase from one round to the next, or that Alg.~\ref{alg:arf} fail to terminate altogether for some datasets. (In practice, this behavior is mitigated by increasing $\delta$ or setting some maximum number of iterations.) For instance, on MNIST, we generally find accuracy plateauing around 65\% after five rounds with little improvement thereafter. 
Of course, the same caveats apply to any asymptotic guarantee. Finite sample results are rare in the RF literature, although there has been some recent work in this area \citep{gao2020}. 

Another potential difficulty for our approach is selecting an optimal density estimation subroutine. KDE relies on a smoothness assumption (A4), while MLE requires a (local) parametric model. 
Bayesian inference imposes a prior distribution, which may bias results. 
All three methods will struggle when their assumptions are violated.
Resampling alternatives such as permutations or bootstrapping do not produce any data that was not observed in the training set and may therefore raise privacy concerns. No approach is generally guaranteed to strike the optimal balance between efficiency, accuracy, and privacy, and so the choice of which combination of methods to employ is irreducibly context-dependent.

We emphasize that our method performs well in a range of settings without any model tuning. However, we acknowledge that optimal performance likely depends on RF parameters \citep{scornet_tuning, probst2019}. In particular, there is an inherent trade-off between the goals of minimizing errors of density ($\epsilon_2$) and errors of convergence ($\epsilon_3$) in finite samples. Grow trees too deep, and leaves will not contain enough data to accurately estimate marginal densities; grow trees too shallow, and ARFs may not satisfy the local independence criterion. Meanwhile, the $\texttt{mtry}$ parameter has been shown to control sparsity in low signal-to-noise regimes \citep{mentch2020}. Smaller values may therefore be appropriate when $\epsilon_3$ is large, in order to regularize the forest. 
Adding more trees tends to improve density estimates, though this incurs extra computational cost in both time and memory \citep{probst2017}. 
Despite these considerations, we reiterate that ARFs do remarkably well with default parameters.

The ethical implications of generative models are potentially fraught. Deepfakes have attracted particular attention in this regard \citep{DeRuiter2021, Ohman2020, deepfakes_elections}, as they can deceive their audience into believing that people said or did things they never in fact said or did. These dangers are most acute with convolutional neural networks or other architectures optimized for visual and audio data. Despite and in full awareness of these concerns, we point out that generative models also present a valuable ethical opportunity, since they may preserve the privacy of data subjects by creating datasets that preserve statistical signals without exposing the personal information of individuals. However, the privacy-utility trade-off can be unpredictable with synthetic data \citep{stadler2022}. As with all powerful technologies, caution is advised and regulatory frameworks are welcome.

\section{CONCLUSION}\label{sec:conclusion}

We have introduced a novel procedure for learning joint densities and generating synthetic data using a recursive, adversarial variant of unsupervised random forests. The method is provably consistent under reasonable assumptions, and performs well in experiments on simulated and real-world examples. Our $\textsc{Forde}$ algorithm is more accurate than other CART-based density estimators and compares favorably to leading PC algorithms. Our $\textsc{Forge}$ algorithm is competitive with deep learning models for data generation on tabular data benchmarks, and routinely executes some 100 times faster. An $\texttt{R}$ package, $\texttt{arf}$, is available on $\texttt{CRAN}$. A Python implementation is forthcoming. 

Future work will explore further applications for these methods, such as anomaly detection, clustering, and classification, as well as potential connections with differential privacy \citep{Dwork2008}. 
Though we have focused in this work on unconditional density estimation tasks, it is straightforward to compute arbitrary conditional probabilities with ARFs by reducing the event space to just those leaves that satisfy some logical constraint(s). More complex functionals may be estimated with just a few additional steps---e.g. (conditional) quantiles, CDFs, and copulas---thereby linking these methods with recent work on functional regression with random forests \citep{hothorn2021, Fu2021, cevid2022}.  
Alternative tree-based solutions based on gradient boosting also warrant further exploration, especially given promising recent developments in this area \citep{friedman2020, gao_linCDE}.

\subsection*{Acknowledgments}
MNW and KB received funding from the German Research Foundation (DFG), Emmy Noether Grant 437611051. MNW and JK received funding from the U Bremen Research Alliance/AI Center for Health Care, financially supported by the Federal State of Bremen. 
We are grateful to Cassio de Campos, Gennaro Gala, Robert Peharz, and Alvaro H.C. Correia for their feedback on an earlier draft of this manuscript.

\bibliography{biblio}


\clearpage
\appendix

\thispagestyle{empty}

\onecolumn 

\section{PROOFS}\label{sec:app_proof}
\subsection{Proof of Thm.~\ref{thm:conv}}\label{proof:conv}
To secure the result, we must show that (a) the discriminator reliably converges on the Bayes risk at each iteration $t$; and (b) the generator's sampling strategy drives original and synthetic data closer together, ultimately taking the Bayes risk to $1/2$ as $n, t \rightarrow \infty$. (For the purposes of this proof, we set the tolerance parameter $\delta$ to $0$.)

Take (a) first. This amounts to a consistency requirement for RFs.
The consistency of RF classifiers has been demonstrated under various assumptions about splitting rules and stopping criteria \citep{breiman2004, biau_consistency, biau2010, gao2020}, but these results generally require trees to be grown to purity or even completion (i.e., $n^\ell_b = 1$ for all $\ell, b$). However, this would turn the generator's sampling strategy into a simple copy-paste operation and make intra-leaf density estimation impossible. We therefore follow \citet{Malley2012} in observing that regression procedures constitute probability machines, since $P(Y=1|\mathbf{x}) = \mathbb{E}[Y|\mathbf{x}]$ for $Y \in \{0, 1\}$. 

For simplicity, we focus on the single tree case, as the consistency of the ensemble follows from the consistency of the base method \citep{biau_consistency}. We define $\eta^{(t)}(\mathbf{x}) := P(Y=1|\mathbf{x}, t)$ as the target function for fixed $t$. Let $f_n^{(t)}(\mathbf{x})$ be a tree trained according to (A1)-(A3) on a sample of size $n$ at iteration $t$. Since $L_2$-consistency entails classifier consistency using the soft labeling approach of (A3).(vi), our goal in this section is to show that, for all $t \in \mathbb{N}$, we have:
\begin{align*}
    \lim_{n \rightarrow \infty} \mathbb{E} \Big[ \big(f_n^{(t)}(\mathbf{x}) - \eta^{(t)}(\mathbf{x})\big)^2 \Big] = 0.
\end{align*}
Consistency for RF regression has been established for several variants of the algorithm, occasionally under some constraints on the data generating process \citep{genuer2012, Scornet2015, wager_walther, Biau2016}. Recent work in this area has tended to focus on asymptotic normality \citep{mentch2016, Wager2018, Athey2019, Peng2022}, which requires additional assumptions. These often include an upper bound on leaf sample size, which would complicate our analysis in Thm.~\ref{thm:cons}. To avoid unnecessary difficulties, we borrow selectively from \citet{Meinshausen2006}, \citet{biau_analysis}, \citet{denil2014}, \citet{scornet2016_asymptotics}, and \citet{Wager2018}, striking a delicate balance between theoretical parsimony and fidelity to the classic RF algorithm.\footnote{Several authors have conjectured that RF consistency may not require honesty (A3).(i) or subsampling (A3).(ii) after all. Empirical performance certainly seems unencumbered by these requirements. However, both come with major theoretical advantages---the former by making predictions conditionally independent of the training data while preserving some form of adaptive splits, the latter by avoiding thorny issues arising from duplicated samples when bootstrapping. See \citet[Rmk.~8]{biau_analysis}, \citet[Appx.~B]{Wager2018}, and \citet{tang2018} for a discussion.} 

For full details, we refer readers to the original texts. The main point to recognize is that, under assumptions (A1)-(A3), RFs satisfy the conditions of Stone's theorem \citep{stone1977}, which guarantees the universal consistency of a large family of local averaging methods. 
\citet[Thm.~6.1]{devroye1996} and \citet[Thm.~4.2]{gyorfi2002} show that partitioning estimators such as decision trees qualify provided that (1) $\text{diam}(\mathcal{X}_\ell) \rightarrow_p 0$ and (2) $n_\ell \rightarrow_p \infty$ for all $\ell$ as $n \rightarrow \infty$, effectively creating leaves of infinite density. The former is derived by \citet[Lemma 2]{Meinshausen2006} under (A3).(iii) and (A3).(iv); the latter follows trivially from (A3).(v). Thus RF discriminators weakly converge on the Bayes risk in the large sample limit, completing part (a) of the proof.

Desideratum (b) effectively says that original and synthetic data become indistinguishable as $n$ and $t$ increase. 
Recall that at $t=0$, we generate synthetic data $\tilde{\mathbf{X}}^{(0)} \sim \prod_{j=1}^d P(X_j)$, which becomes input to the discriminator $f^{(0)}_n$. Let $\bm{\theta}^{(0)}$ denote the resulting splits once the discriminator has converged.
In subsequent rounds, synthetic data are sampled according to $\tilde{\mathbf{X}}^{(t+1)} \sim \prod_{j=1}^d P(X_j|\theta_\ell^{(t)})P(\theta_\ell^{(t)})$. (The consistency of coverage estimates is treated separately in Appx.~\ref{proof:l2}.)
We proceed to train a new discriminator and repeat the process. 

Let $P^*$ be the target distribution and $P^{(t)}$ the synthetic distribution at round $t$. For all $t \geq 1$, the input data to the discriminator $f^{(t)}_n$ is the dataset $\mathcal{D}_n^{(t)} \sim 0.5P^* + 0.5P^{(t-1)}$. Our goal in this section is to show that, as $n, t \rightarrow \infty$:
\begin{align*}
    \sup_{\mathbf{x} \in \mathcal{D}_n^{(t)}} |\eta^{(t)}(\mathbf{x}) - 1/2| \rightarrow_p 0.
\end{align*}

An apparent challenge to our recursive strategy for generating synthetic data is posed by self-similar distributions, in which dependencies replicate at ever finer resolutions, as in some fractal equations \citep{mandelbrot1982}. For instance, let $g$ be the Weierstrass function \citep{Weierstrass1895}, and say that $X_2 = g(X_1)$. Then the generative model will tend to produce off-manifold data at each iteration $t$, no matter how small $\text{vol}(\mathcal{X}_\ell)$ becomes. However, this only shows that convergence can fail for finite $t$.
Since the discriminator is consistent, it will accurately identify synthetic points in round $t+1$, pruning the space still further.

Let $[L^{(t)}]$ be the leaves of the discriminator $f^{(t)}_n$, and define the maximum leaf diameter $m_t := \max_{\ell \in [L^{(t)}]} \text{diam}(\mathcal{X}_\ell)$.We say that two samples are \textit{neighbors} in $f_n^{(t)}$ if the model places them in the same leaf.
We show that, as $n, t \rightarrow \infty$, conditional probabilities for neighboring samples converge---including, crucially, original and synthetic counterparts. Our Lipschitz condition (A2) states that for all $\mathbf{x}, \mathbf{x}'$, we have:
\begin{align*}
    |\eta^{(t)}(\mathbf{x}) - \eta^{(t)}(\mathbf{x}')| \leq c_t ~\lVert \mathbf{x} - \mathbf{x}' \rVert_2,
\end{align*}
where $c_t$ denotes the Lipschitz constant at round $t$. 
Suppose that $\mathbf{x}$ and $\mathbf{x}'$ are neighbors. 
Then we can replace the second factor on the rhs with $m_t$, since the $L_2$ distance between neighbors cannot exceed the maximum leaf diameter at round $t$. \citet{Meinshausen2006}'s aforementioned Lemma 2 ensures that this value goes to zero in probability as rounds increase. This could in principle be offset by a sufficient increase in $c_t$ over training rounds, but the second condition of (A2) prevents this, imposing the constraint that $c_t = o(m_t^{-1})$. Thus, for observations in the same leaf, $c_tm_t \rightarrow_p 0$ as $t \rightarrow \infty$. 
Because original and synthetic samples are equinumerous in all leaves following the generative step, each original sample has a synthetic counterpart to which it is arbitrarily close in $L_2$ space as $t$ grows large. Since no feature values are sufficient to distinguish between the two classes in any region, all conditional probabilities go to 1/2, and Bayes risk therefore also converges to 1/2 in probability. This concludes the proof.

\subsection{Proof of Lemma~\ref{lm:decomp}} 
Define the first-order approximation to $p$ satisfying local independence:
\begin{equation*}\label{eq:fir}
    \hat{p}(\mathbf{x}) := \frac{1}{B} \sum_{\ell, b: \mathbf{x} \in \mathcal{X}^\ell_b} ~p(\theta_b^\ell) ~\prod_{j=1}^d p(x_j|\theta^\ell_b).
\end{equation*}
We also define the root integrated squared error (RISE), i.e. the Euclidean distance between probability densities:
\begin{equation*}
    \text{RISE}(p, q) := \bigg( \int_{\mathcal{X}} \Big( p(\mathbf{x}) - q(\mathbf{x}) \Big)^2 ~d\mathbf{x} \bigg)^{1/2}.
\end{equation*}
By the triangle inequality, we have:
\begin{align*}
    \text{RISE}(p, q) &\leq \text{RISE}(p, \hat{p}) + \text{RISE}(\hat{p}, q). 
\end{align*}
Squaring both sides, we get:
\begin{align*}
    \text{ISE}(p, q) &\leq \text{ISE}(p, \hat{p}) + \text{ISE}(\hat{p}, q) + 2  ~\text{RISE}(p, \hat{p}) ~\text{RISE}(\hat{p}, q). 
\end{align*}
Adding a nonnegative value to the rhs, we can reduce the expression:
\begin{align*}
    \text{ISE}(p, q) &\leq \text{ISE}(p, \hat{p}) + \text{ISE}(\hat{p}, q) + 2  ~\text{RISE}(p, \hat{p}) ~\text{RISE}(\hat{p}, q) + \big(\text{RISE}(p, \hat{p}) - \text{RISE}(\hat{p}, q) \big)^2 \\
    &= 2 \big( \text{ISE}(p, \hat{p}) + \text{ISE}(\hat{p}, q) \big).
\end{align*}
Now observe that we can rewrite both ISE formulae in terms of our predefined residuals (Eqs. \ref{eq:eps1}-\ref{eq:eps3}):
\begin{align*}
    \text{ISE}(p, \hat{p}) &= \int_\mathcal{X} \bigg(\frac{1}{B} \sum_{\ell, b: \mathbf{x} \in \mathcal{X}^\ell_b} \Big( p(\mathbf{x}|\theta^\ell_b)~p(\theta^\ell_b) - \prod_{j=1}^d p(x_j|\theta^\ell) ~p(\theta^\ell_b) \Big) \bigg)^2 ~d\mathbf{x} \\
    &= \frac{1}{B^2}\int_\mathcal{X} \bigg( \sum_{\ell, b: \mathbf{x} \in \mathcal{X}^\ell_b} p(\theta^\ell_b) \epsilon_3 \bigg)^2 ~d\mathbf{x}. \\
    \text{ISE}(\hat{p}, q) &= \int_\mathcal{X} \bigg(\frac{1}{B} \sum_{\ell, b: \mathbf{x} \in \mathcal{X}^\ell_b} \Big( \prod_{j=1}^d p(x_j|\theta^\ell_b) ~p(\theta^\ell_b) - \prod_{j=1}^d q(x_j; \bm{\theta}_{b,j}^\ell) ~q(\theta^\ell_b) \Big) \bigg)^2 ~d\mathbf{x} \\
    &= \frac{1}{B^2}\int_\mathcal{X} \bigg( \sum_{\ell, b: \mathbf{x} \in \mathcal{X}^\ell_b} \Big( p(\theta^\ell_b)\epsilon_2 + \epsilon_1\prod_{j=1}^d p(x_j|\theta^\ell_b) - \epsilon_1 \epsilon_2 \Big) \bigg)^2 ~d\mathbf{x}.
\end{align*}
We replace the interior squared terms for ease of presentation:
\begin{align*}
    \alpha &:= \sum_{\ell, b: \mathbf{x} \in \mathcal{X}_b^\ell} p(\theta^\ell_b) \epsilon_3 \\
    \beta &:= \sum_{\ell, b: \mathbf{x} \in \mathcal{X}_b^\ell} \Big( p(\theta^\ell_b)\epsilon_2 + \epsilon_1\prod_{j=1}^d p(x_j|\theta^\ell_b) - \epsilon_1 \epsilon_2 \Big).
\end{align*}
Finally, we take expectations on both sides:
\begin{align*}
    \text{MISE}(p, q) \leq 2B^{-2}~\mathbb{E} \Bigg[ \int_\mathcal{X} \alpha^2 + \beta^2 ~d\mathbf{x} \Bigg],
\end{align*}
where we have exploited the linearity of expectation to pull the factor outside of the bracketed term, and the monotonicity of expectation to preserve the inequality.

\subsection{Proof of Theorem~\ref{thm:cons}}\label{proof:l2}
Lemma~\ref{lm:decomp} states that error is bounded by a quadratic function of $\epsilon_1, \epsilon_2, \epsilon_3$. Thus for $L_2$-consistency, it suffices to show that $\mathbb{E}[\epsilon_j^2] \rightarrow 0$, for $j \in \{1, 2, 3\}$. Since this is already established by Thm.~\ref{thm:conv} for $j=3$, we focus here on errors of coverage and density. Start with $\epsilon_1$. A general version of the Glivenko-Cantelli theorem \citep{Vapnik2015} guarantees uniform convergence of empirical proportions to population proportions. Let $\mathcal{L}$ denote the set of all possible hyperrectangular subspaces induced by axis-aligned splits on $\mathcal{X}$. Then the following holds with probability 1: 
\begin{align*}
    \lim_{n \rightarrow \infty} \sup_{\ell \in \mathcal{L}} \big|p(\theta^\ell) - q_n(\theta^\ell)\big| = 0.
\end{align*}

Next, take $\epsilon_2$. (A4) guarantees that $p$ satisfies the consistency conditions for univariate KDE \citep{silverman1986, wand1994, Gramacki_kde}, while condition (v) of (A3) ensures that within-leaf sample size increases even as leaf volume goes to zero \citep[Lemma 2]{Meinshausen2006}. Our kernel is a nonnegative function $K: \mathbb{R}^d \rightarrow \mathbb{R}$ that integrates to 1, parametrized by the bandwidth $h$:
\begin{align*}
    p_h(\mathbf{x}) = \frac{1}{nh} \sum_{i=1}^n K\Big(\frac{\mathbf{x} - \mathbf{x}_i}{h}\Big).
\end{align*}
Using standard arguments, we take a Taylor series expansion of the MISE and minimize the \textit{asymptotic} MISE (AMISE):
\begin{equation*}
    \text{AMISE}(p, p_h) = \frac{1}{nh}R(K) + \frac{1}{4} h^4 \mu_2(K)^2R(p''),
\end{equation*}
where
\begin{align*}
    R(K) &= \int K(x)^2 ~dx, \\
    \mu_2(K) &= \int x^2 K(x) ~dx, ~\text{and}\\
    R(p'') &= \int p''(x)^2 ~dx.
\end{align*}
For example values of these variables under specific kernels, see \citep[Appx. B]{wand1994}. Under (A4), it can be shown that 
\begin{equation*}
    \text{MISE}(p, p_h) = \text{AMISE}(p, p_h) + o\big((nh)^{-1} + h^4\big).
\end{equation*}
Thus if $(nh)^{-1} \rightarrow 0$ and $h \rightarrow 0$ as $n \rightarrow \infty$, the asymptotic approximation is exact and $\mathbb{E}[\epsilon_2^2] \rightarrow 0$.  

These results, combined with the proof of Thm.~\ref{thm:conv} (see Appx.~\ref{proof:conv}), establish that errors of coverage, density, and convergence all vanish in the limit. Thus $\mathbb{E}[\epsilon_j^2] \rightarrow 0$ for $j \in \{1, 2, 3\}$, and the proof is complete.

\newpage
\section{EXPERIMENTS}\label{sec:app_exp}
Our experiments do not include any personal data, as defined in Article 4(1) of the European Union's General Data Protection Regulation. All data are either simulated or from publicly available resources. 
We performed all experiments on a dedicated 64-bit Linux platform running Ubuntu 20.04 with an AMD Ryzen Threadripper 3960X (24 cores, 48 threads) CPU, 256 gigabyte RAM and two NVIDIA Titan RTX GPUs. We used $\texttt{R}$ version 4.1.2 and Python version 3.7.12. Further details on the environment setup are provided in the supplemental code. 

\subsection{Simulations}
The $\texttt{cassini}$, $\texttt{shapes}$, and $\texttt{smiley}$ simulations are all available in the $\texttt{mlbench}$ $\texttt{R}$ package; the $\texttt{twomoons}$ problem is available in the $\texttt{fdm2id}$ $\texttt{R}$ package. Default parameters were used throughout, with fixed sample size $n=2000$. 

\subsection{Twenty Datasets}\label{sec:appx_pc}

The Twenty Datasets benchmark was originally proposed by \citet{Davis_2021}. A conventional training/validation/test split is widely used in the PC literature. Because our method does not include any hyperparameter search, we combine training and validation sets into a single training set. We downloaded the data from \url{https://github.com/joshuacnf/Probabilistic-Generating-Circuits/tree/main/data} and include the directory in our project GitHub repository for completeness. 
All datasets are Boolean, with sample size and dimensionality given in Table \ref{tab:20datasets}.

\begin{table}[htbp]
\scriptsize
\centering
\caption{Summary of datasets included in the Twenty Datasets benchmark.}
\label{tab:20datasets}
\begin{tabular}{lrrrr}
\toprule
Dataset   & Train  & Validation & Test  & Dimensions \\ 
\midrule
\texttt{nltcs}     & 16181  & 2157       & 3236  & 16             \\
\texttt{msnbc}     & 291326 & 38843      & 58265 & 17             \\
\texttt{kdd}       & 180092 & 19907      & 34955 & 64             \\
\texttt{plants}    & 17412  & 2321       & 3482  & 69             \\
\texttt{audio}     & 15000  & 2000       & 3000  & 100            \\
\texttt{jester}    & 9000   & 1000       & 4116  & 100            \\
\texttt{netflix}   & 15000  & 2000       & 3000  & 100            \\
\texttt{accidents} & 12758  & 1700       & 2551  & 111            \\
\texttt{retail}    & 22041  & 2938       & 4408  & 135            \\
\texttt{pumsb}     & 12262  & 1635       & 2452  & 163            \\
\texttt{dna}       & 1600   & 400        & 1186  & 180            \\
\texttt{kosarek}   & 33375  & 4450       & 6675  & 190            \\
\texttt{msweb}     & 29441  & 3270       & 5000  & 294            \\
\texttt{book}      & 8700   & 1159       & 1739  & 500            \\
\texttt{movie}     & 4524   & 1002       & 591   & 500            \\
\texttt{webkb}     & 2803   & 558        & 838   & 839            \\
\texttt{reuters}   & 6532   & 1028       & 1540  & 889            \\
\texttt{20ng}      & 11293  & 3764       & 3764  & 910            \\
\texttt{bbc}       & 1670   & 225        & 330   & 1058           \\
\texttt{ad}        & 2461   & 327        & 491   & 1556           \\
\bottomrule
\end{tabular}
\end{table}

Results for competitors are reported in the cited papers:
\begin{itemize}[noitemsep]
    \item Einsum networks \citep{peharz2020}
    \item Random sum-product networks \citep{ratspn}
    \item Probabilistic generating circuits \citep{zhang2021}
    \item Strudel \citep{dang2022}
    \item Continuous mixtures of Chow-Liu trees \citep{correia2022}.
\end{itemize}

\subsection{Tabular GANs}\label{sec:appx_gan}
For benchmarking generative models on real-world data, we use the benchmarking pipeline proposed by \citet{Xu2019}. In detail, the workflow is as follows:
\begin{enumerate}
    \item Load classification datasets used in \citet{Xu2019}, namely $\texttt{adult}$, $\texttt{census}$, $\texttt{credit}$, $\texttt{covertype}$, $\texttt{intrusion}$, $\texttt{mnist12}$, and $\texttt{mnist28}$. Note that the type of prediction task does not affect the process of synthetic data generation, so we omit the single regression example ($\texttt{news}$) for greater consistency. 
    \item Split the data into training and test sets (see Table~\ref{benchmark_details} for details).
    \item Train the generative models $\textsc{Forge}$ (number of trees $=10$, minimum node size $=5$),  $\textsc{ctgan}$\footnote{\url{https://sdv.dev/SDV/api_reference/tabular/ctgan.html}. MIT License.} (batch size $= 500$, epochs $=300$), $\textsc{tvae}$\footnote{\url{https://sdv.dev/SDV/api_reference/tabular/tvae.html}. MIT License.} (batch size $= 500$, epochs $=300$), $\textsc{ctab-gan+}$\footnote{\url{https://github.com/Team-TUD/CTAB-GAN-Plus}} (batch size $= 500$, epochs $= 150$), $\textsc{it-gan}$\footnote{\url{https://github.com/leejaehoon2016/ITGAN}. Samsung SDS Public License V1.0.} (batch size $= 2000$, epochs $= 300$) and $\textsc{rcc-gan}$\footnote{\url{https://github.com/EsmaeilpourMohammad/RccGAN}} (batch size $= 500$, epochs $= 300$).
    \item Generate a synthetic dataset of the same size as the training set using each of the generative models trained in step (3), measuring the wall time needed to execute this task.
    \item Train a set of supervised learning algorithms (see Table~\ref{benchmark_details} for details): (a) on the real training data set (i.e., the \textit{Oracle}); and (b) on the synthetic training datasets generated by $\textsc{Forge}$,  $\textsc{ctgan}$, $\textsc{tvae}$, $\textsc{ctab-gan+}$ and $\textsc{rcc-gan}$.
    \item Evaluate the performance of the learning algorithms from step (5) on the test set.
    \item For each dataset, average performance metrics (accuracy, F1-scores) across learners. We report F1-scores for the positive class, e.g. `>50k' for \texttt{adult}, `+50000' for \texttt{census} and `1' for \texttt{credit}.
\end{enumerate}

\begin{table}[htbp]
\scriptsize
\centering
\caption{Benchmark Setup. Supervised learning algorithms for prediction: (A) Adaboost, estimators = 50 , (B) Decision Tree, tree depth for binary/multiclass target = 15/30, (C) Logistic Regression, (D) MLP, hidden layers for binary/multiclass target = 50/100}
\label{benchmark_details}
\begin{tabular}{llll}
\toprule
Dataset & Train/Test & Learner & Link to dataset\\ \midrule
\texttt{adult} \citep{Dua2019} & 23k/10k & A,B,C,D &\url{http://archive.ics.uci.edu/ml/datasets/adult}\\ 
\texttt{census} \citep{Dua2019} & 200k/100k & A,B,D &\url{https://archive.ics.uci.edu/ml/datasets/census+income} \\
\texttt{covertype} \citep{covertype} & 481k/100k & A,D &\url{https://archive.ics.uci.edu/ml/datasets/covertype} \\
\texttt{credit} \citep{credit} & 264k/20k & A,B,D &\url{https://www.kaggle.com/mlg-ulb/creditcardfraud}\\
\texttt{intrusion} \citep{Dua2019} & 394k/100k & A,D &\url{http://archive.ics.uci.edu/ml/datasets/kdd+cup+1999+data}\\
\texttt{mnist12} \citep{mnist} & 60k/10k & A,D & \url{http://yann.lecun.com/exdb/mnist/index.html}\\
\texttt{mnist28} \citep{mnist} & 60k/10k & A,D & \url{http://yann.lecun.com/exdb/mnist/index.html}\\
\bottomrule 
\end{tabular}\end{table} 

\subsection{Run Time}\label{app:runtime}

\begin{figure}[htbp]
  \centering
  \includegraphics[width=0.85\textwidth]{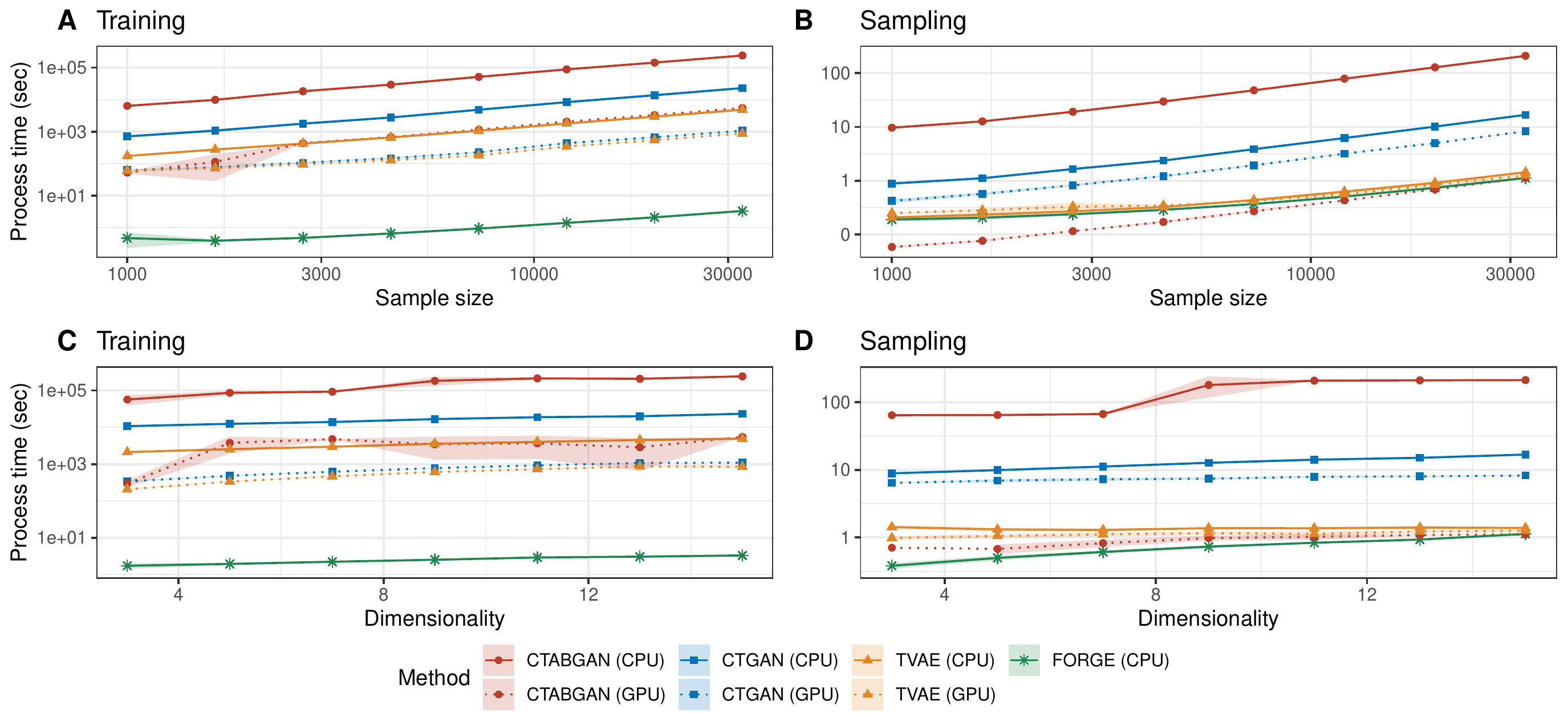}
  \caption{Complexity curves. \textbf{(A)}: Processing time as a function of sample size, using stratified subsamples of the \texttt{adult} dataset. \textbf{(B)}: Processing time as a function of dimensionality, using random features from the \texttt{adult} dataset.}
  \label{fig:process_time}
\end{figure}

In order to evaluate the run time efficiency of $\textsc{Forge}$, we chose to focus on the smallest dataset of the benchmark study in Sect.~\ref{sec:benchmark}, namely \texttt{adult}. We \textbf{(A)} drew stratified subsamples and \textbf{(B)} drew covariate subsets. For step \textbf{(B)}, the target variable is always included. We select an equal number of continuous/categorical covariates when possible and use all $n = 32,561$ instances. Results in terms of processing time are visualized in Fig.~\ref{fig:process_time}. 

\subsection{Image Data}\label{sec:mnist}
We include results on the $\texttt{mnist12}$ and $\texttt{mnist28}$ datasets here, both included in the original \citet{Xu2019} pipeline. Benchmarking against $\textsc{ctgan}$ and $\textsc{tvae}$ (other methods proved too slow to test), we find that $\textsc{Forge}$ outperforms both competitors in accuracy, F1-score, and speed (see Table \ref{tbl:mnist}).

\begin{table}[htbp]
\scriptsize
\centering
\caption{Performance on $\texttt{mnist}$ datasets from the \citet{Xu2019} benchmark for $\textsc{ctgan}$ and $\textsc{tvae}$ vs. $\textsc{Forge}$ We report average results across five replicates $\pm$ the associated standard error. Winning results in bold.}
\label{tbl:mnist}
\begin{tabular}{llccr}
  \toprule
Dataset & Model & Accuracy $\pm$ SE  & F1 $\pm$ SE & Time (sec) \\ 
  \midrule
  \texttt{mnist12} & \textit{Oracle} & \textit{0.892 $\pm$ 0.003} & \textit{0.891 $\pm$ 0.003} &  \\ 
   classes = 10 & $\textsc{Forge}$ & \textbf{0.799} $\pm$ 0.007 & \textbf{0.795} $\pm$ 0.007 & \textbf{32.3} \\ 
   $n $ = 70,000 & $\textsc{ctgan}$ & 0.172 $\pm$ 0.032 & 0.138 $\pm$ 0.032 & 2737.4 \\ 
   $d$ = 144 & $\textsc{tvae}$ & 0.763 $\pm$ 0.002 & 0.761 $\pm$ 0.002 & 1143.8 \\ \midrule
  \texttt{mnist28} & \textit{Oracle} & \textit{0.918 $\pm$ 0.002} & \textit{0.917 $\pm$ 0.002} & \\ 
   classes = 10 & $\textsc{Forge}$ & \textbf{0.729} $\pm$ 0.008 & \textbf{0.723} $\pm$ 0.008 & \textbf{169.5} \\ 
   $n $ = 70,000 & $\textsc{ctgan}$ & 0.197 $\pm$ 0.051 & 0.167 $\pm$ 0.051 & 14415.4 \\ 
   $d$ = 784 & $\textsc{tvae}$ & 0.698 $\pm$ 0.016 & 0.697 $\pm$ 0.016 & 5056.0 \\ 
   \bottomrule
\end{tabular}
\end{table}

\begin{figure}[htbp]
  \centering
  \includegraphics[width=\textwidth]{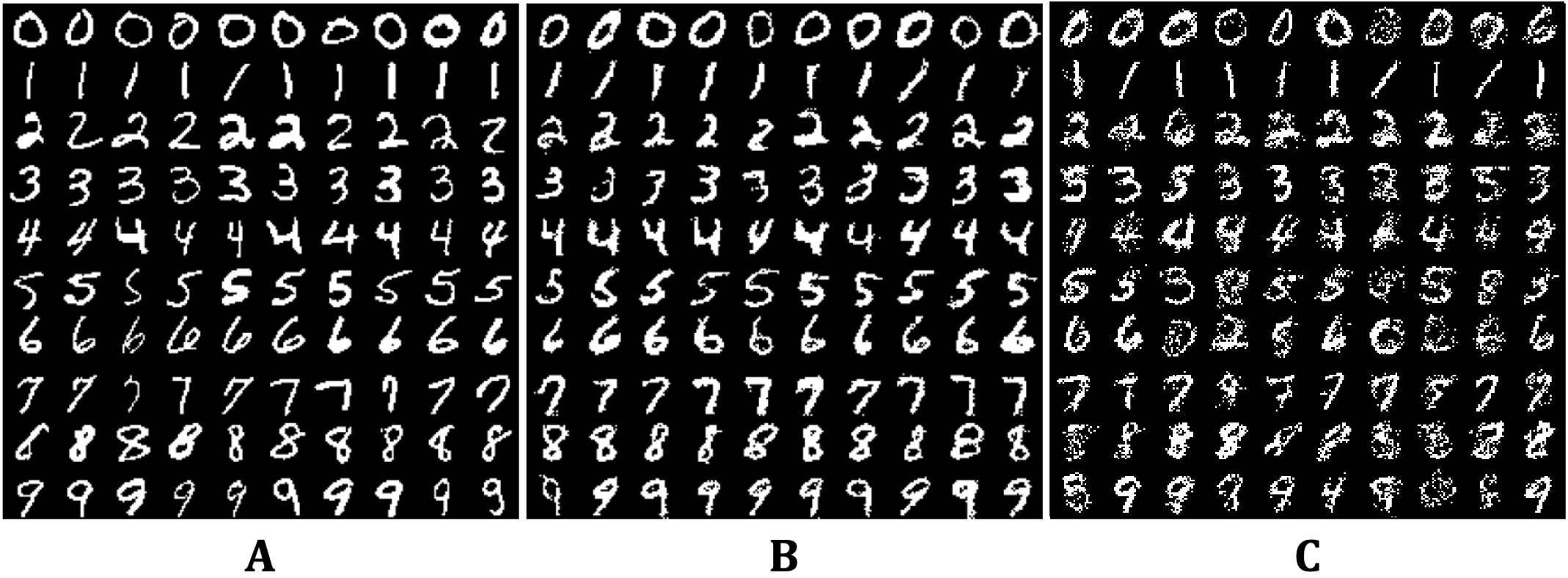}
  \caption{Results from $\texttt{mnist28}$ experiment. \textbf{(A)}: Original samples. \textbf{(B)}: Samples generated by cGAN. \textbf{(C)}: Samples generated by $\textsc{Forge}$.}
  \label{fig:mnist28}
\end{figure}

However, since MNIST is not a tabular data problem, perhaps a more relevant comparison would be against convolutional networks specifically designed for image data. We train a conditional GAN with convolutional layers \citep{cgan2014} and find that the resulting $\texttt{cGAN}$ clearly outperforms $\textsc{Forge}$ (see Fig. \ref{fig:mnist28}). This result is expected, given that our method is not optimized for image data. It also illustrates a limitation of our approach, which excels in speed and flexibility but is no match for deep learning methods on structured datasets.

\end{document}